\documentclass[letterpaper, 10 pt, conference]{IEEEtran}

\usepackage{amsthm}
\usepackage[numbers,sort&compress]{natbib}
\usepackage{multicol}
\usepackage[bookmarks=true]{hyperref}
\usepackage{amsmath}
\usepackage{amssymb}
\usepackage{graphicx}
\usepackage{xcolor}
\usepackage{comment}
\usepackage[normalem]{ulem}
\usepackage[algo2e,ruled,vlined,linesnumbered,resetcount,norelsize]{algorithm2e}
\usepackage{algorithm}
\usepackage{booktabs}

\usepackage{enumitem}

\usepackage{booktabs}
\usepackage{diagbox}

\usepackage{fontawesome5}

\newtheorem{remark}{Remark}

\newcommand{\marginXW}[1]{\marginpar{\color{purple}\tiny\ttfamily#1}}
\newcommand{\XW}[1]{{\color{purple}#1}}

\begin{document}

\title{\LARGE \bf CARoL: Context-aware Adaptation for Robot Learning}

\author{Zechen Hu, Tong Xu, Xuesu Xiao, and Xuan Wang}



%

\maketitle

\begin{abstract}
    Using Reinforcement Learning (RL) to learn new robotic tasks from scratch is often inefficient. Leveraging prior knowledge has the potential to significantly enhance learning efficiency, which, however, raises two critical challenges: how to determine the relevancy of existing knowledge and how to adaptively integrate them into learning a new task. In this paper, we propose \textit{Context-aware Adaptation for Robot Learning} (CARoL) \footnote{Code will be publicly available on \faGithub.}, a novel framework to efficiently learn a similar but distinct new task from prior knowledge. CARoL incorporates context awareness by analyzing state transitions in system dynamics to identify similarities between the new task and prior knowledge. It then utilizes these identified similarities to prioritize and adapt specific knowledge pieces for the new task. Additionally, CARoL has a broad applicability spanning policy-based, value-based, and actor-critic RL algorithms. 
    We validate the efficiency and generalizability of CARoL on both simulated robotic platforms and physical ground vehicles. 
    The simulations include CarRacing and LunarLander environments, where CARoL demonstrates faster convergence and higher rewards when learning policies for new tasks. 
    In real-world experiments, we show that CARoL enables a ground vehicle to quickly and efficiently adapt policies learned in simulation to smoothly traverse real-world off-road terrain.

\end{abstract}

\IEEEpeerreviewmaketitle

\section{Introduction}
In recent years, Reinforcement Learning (RL) approaches have achieved remarkable success in advanced robotic control and complex task learning in dynamic environments, enabling applications across various domains, such as autonomous navigation~\cite{xiao2022motion, xu2023benchmarking}, 
manipulation~\cite{sun2022fully, zhu2020ingredients}, 
and human-robot interaction~\cite{roveda2020model}. 
Despite these advancements, RL methods are typically computationally demanding, as they rely on repeated trial-and-error exploration to discover high-reward outcomes. 

\begin{figure}
    \centering
    \includegraphics[width=0.4\textwidth]{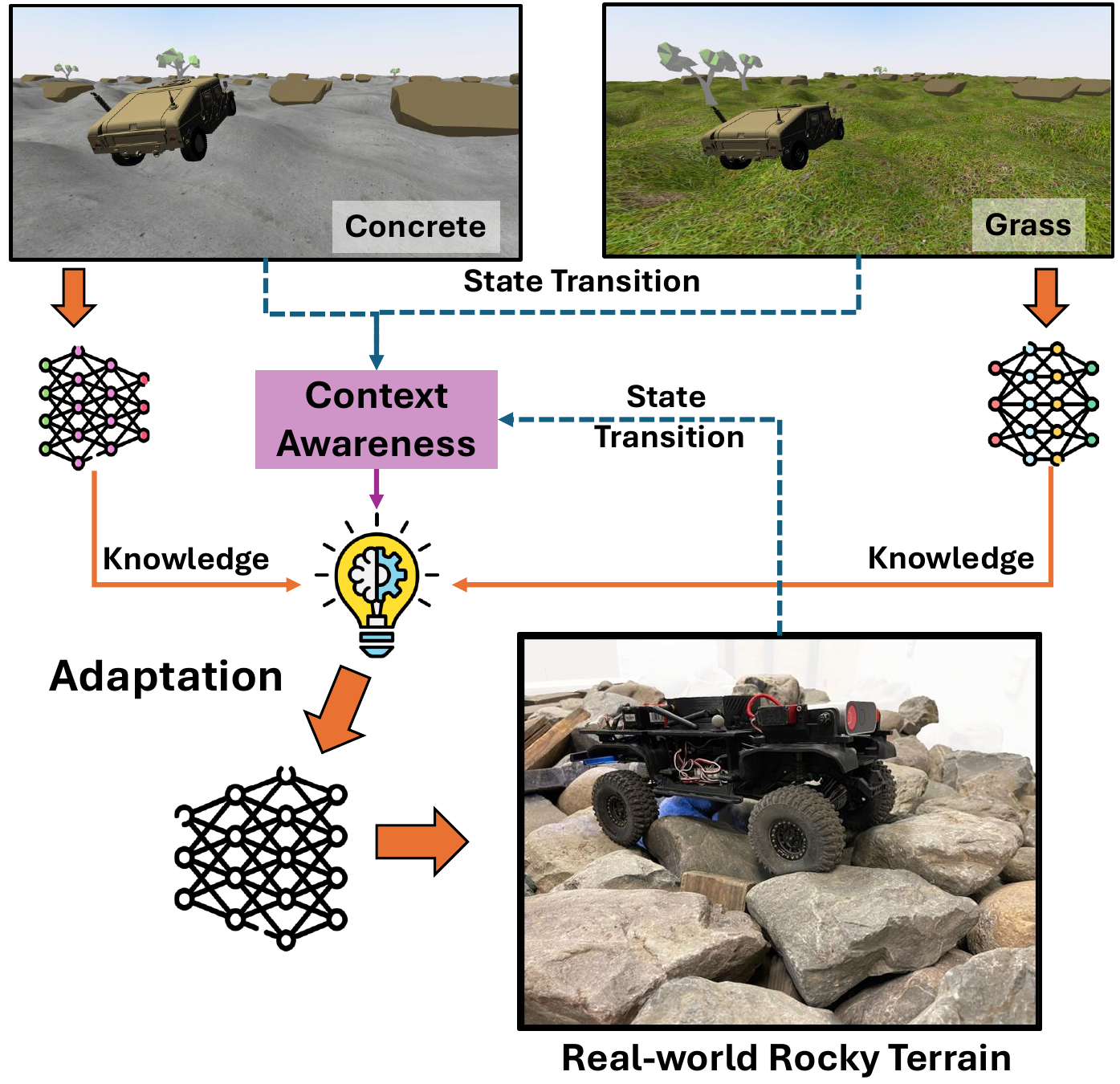}
    \caption{An example of autonomous off-road navigation involves a vehicle extensively trained in (simulated) grass and concrete environments. However, its next task requires it to navigate on (real-world) unseen rocky terrain. To accomplish this, the vehicle must effectively leverage the prior knowledge learned from previous tasks to solve this new challenge.}
    \label{fig:intro}
\end{figure}

Knowledge fusion~\cite{chiu2024flexible} and adaptation~\cite{ruder2017knowledge, xiao2024safe} provide promising approaches to address the inefficiency of RL.
They leverage knowledge (such as a learned control policy, approximated value function, etc.) from previously explored tasks to accelerate training on new tasks, eliminating the need to train from scratch for every scenario.
For example, consider a vehicle navigating highly complex off-road terrain as shown in Fig. \ref{fig:intro}. Suppose the vehicle has undergone extensive training in several existing environments, it should ideally be capable of adapting to a new type of terrain by utilizing previously learned knowledge.

Our goal is to leverage prior knowledge to enhance learning efficiency of a new task. Existing indiscriminate knowledge fusion methods~\cite{wang2021selective} do not consider (and exploit) the relationship between previous and new environments, i.e., being unaware of the environment context, even though many skills are context-specific. Instead, robots should selectively prioritize and utilize knowledge/skills that are relevant to the new task. This poses two key challenges: how to effectively quantify the relevance of existing knowledge to a new task and how to adaptively integrate it into the learning process.

In this work, we propose \textit{\textbf{C}ontext-aware \textbf{A}daptation for \textbf{Ro}bot \textbf{L}earning} (CARoL), a framework that leverages context awareness to enable efficient adaptation to new tasks using prior knowledge. 
CARoL introduces a novel approach to quantify task similarities by using \textit{state transition} representations as contextual indicators, which is different from other context-awareness works that commonly use environment~\cite{kumar2021rma, lee2020learning} or robot~\cite{yang2020multi, peng2018deepmimic} representations.
The rationale behind this choice is that the state transition is the core of a Markov Decision Process (MDP) upon which reinforcement learning is established. State transition inherently captures the combined effects of the environment and the robot, thus, is a more straightforward and comprehensive measure of task similarity, without the need of carefully choosing features from environments or robots. 
For instance, steep inclines and partially icy surfaces may differ significantly in their environmental characteristics but can produce similar state transitions such as wheel traction loss. By leveraging this shared context, control policies can be adapted in a more general and robust manner.

CARoL consists of three steps. In the first step, we focus on obtaining prior knowledge in existing environments. Unlike traditional multi-task learning, which focuses solely on training teacher policies for different tasks, our method additionally trains a state transition representation to capture the `context' of each task. 
The second step involves evaluating contextual similarities by comparing each state transition representation with the state transition of the new task. Finally, we use the similarity to prioritize the most contextually appropriate knowledge for adaption, enabling efficient learning for the new task. Such adaption applies to policy-based, value-based and actor-critic RL algorithms.
To summarize, the contributions of CARoL are threefold:
\begin{itemize}
    \item CARoL explicitly leverages state transition to identify task-specific contextual similarity, enabling more targeted learning rather than indiscriminate fusion/adaptation of prior knowledge.
    \item CARoL has broad applicability that can be integrated into both policy-based and value-based (or combined actor-critic) RL algorithms to achieve knowledge adaptation.
    \item Our experimental study demonstrates the effectiveness of CARoL, not only in simulation environments but also through validation on physical off-road navigation tasks.
\end{itemize}

\section{Related Work} \label{sec:literature}
Leveraging prior knowledge is closely related to knowledge fusion, such as multi-task reinforcement learning. 
Learning adaptation enhances efficiency when tackling new tasks.
Additionally, evaluating the relevance of existing knowledge to a new task is closely linked to context-awareness. In this section, we provide a detailed review of the relevant literature.






\subsection{Multi-task Reinforcement Learning}
By leveraging shared knowledge, multi-task learning seeks to enable simultaneous learning of multiple tasks, thus, improving sample efficiency and generalization in robotics and RL~\cite{zhang2021survey}. For example, \citet{wilson2007multi} introduced a hierarchical Bayesian framework for multi-task RL to improve knowledge transfer by exploiting shared structures among related tasks. 
Building on this, \citet{hausman2018learning} further enabled robots to acquire transferable skills across a variety of tasks through learning a shared embedding space. MT-Opt~\cite{kalashnikov2021mt} demonstrated how robots can learn a broad spectrum of skills. For quadrotor control tasks, \citet{xing2024multi} leveraged shared physical dynamics to enhance sample efficiency and task performance. 
Although parameter sharing across tasks can intuitively improve data efficiency, during the training process, gradients from different tasks can interfere negatively with each other. To mitigate this, methods such as policy distillation \cite{rusu2015policy, teh2017distral} and actor-mimic \cite{parisotto2015actor} are proposed to train a single policy by using the guidance of several expert teachers. These methods perform well on tasks that have already been encountered or on tasks with known structures. However, in most robotic applications, robots often face unknown tasks, 
making it difficult to ensure the performance of fused control strategies in these unfamiliar scenarios.

\subsection{Learning Adaptation for Robotics}
To guarantee performance on new tasks, knowledge fusion typically requires an additional adaptation process. To this end, various approaches have been proposed. Lifelong learning, for example, maintains a single model that evolves over time to handle all encountered tasks. This approach enables robots to continuously acquire, adapt, and transfer knowledge throughout their operational lifespan. For instance, \citet{liu2021lifelong} leveraged lifelong learning techniques, specifically Gradient Episodic Memory~\cite{lopez2017gradient}, to learn a navigation policy complementary to classical motion planners to adapt to new navigation scenarios without forgetting previous ones.  \citet{liu2019lifelong} built on federated learning to efficiently acquire prior knowledge from different environments and dynamically adjust the shared model to support robots in quickly adapting to diverse navigation tasks. Similarly, ~\citet{logacjov2021learning} adopted dynamic generation and adjustment of network structures to store and adapt prior knowledge, leveraging unsupervised knowledge integration for new task adaptation in long-term human-robot interactions with a simulated humanoid robot.
With the development of Large Language Models (LLMs), \citet{parakh2024lifelong} used LLMs as planners that decompose high-level instructions into executable steps, enabling robots to adapt to new tasks with human assistance.
Lifelong learning is highly storage-efficient, as it retains only one model that solves all scenarios over the learning lifespan. This also makes it susceptible to task order bias and catastrophic forgetting. Additionally, while lifelong learning tries to avoid forgetting existing knowledge, it may negatively impact the efficiency of acquiring a new one due to contradicting gradients.

Meta-learning~\cite{kaushik2020fast}, on the other hand, adopts a different adaptation mechanism. It trains a meta-model with a set of initial parameters derived from prior knowledge. This meta-model can then be adapted to many other models, one for each task.
Notable examples for metal-learning include applications in legged robots for adapting to changing dynamics~\cite{song2020rapidly} and optimization-based methods like Model-Agnostic Meta-Learning (MAML)~\cite{finn2017model}, which learns initializations that allow fast policy convergence on new tasks.
In terms of fusing prior knowledge, meta-models typically treat all prior knowledge equally, which lacks the capability to dynamically prioritize specific prior knowledge relevant to the context of a new task.

\subsection{Context-awareness for Reinforcement Learning}
Context-aware learning considers a set of Markov Decision Processes (MDPs) that share the same state and action spaces but differ in transition probabilities and rewards based on contextual variations. Thus, a learning model can be made adaptive, by using context-awareness as an augmented input~\cite{hallak2015contextual,li2018context,lee2020context,kumar2021rma,yu2017preparing,liang2023context}. For example, in the work by~\citet{hallak2015contextual}, contextual MPD is proposed to model how human decision-making varies depending on their surrounding environment. Context-Aware Policy reuSe (CAPS) \cite{li2018context} leverages contexts as identifiers to determine when and which source policies should be reused.
For robotic applications, \citet{lee2020context} used a latent context vector to capture robot-specific structural features, enabling the prediction of future states.
For quadruped robot control on diverse terrains, ~\citet{kumar2021rma} and \citet{peng2020learning} treated environmental encodings as contexts, which are then used as an extra argument to a base policy for quick adaption. Specifically, \citet{peng2020learning} directly derived environmental encodings, while \citet{kumar2021rma} employed an additional mechanism to infer these encodings inversely from state transitions. For these existing works, context is typically represented as latent encodings of either environments or robots, generated during end-to-end training processes. In contrast, our work directly uses state transitions as contexts, bypassing the need for latent representations.
Moreover, while existing methods primarily rely on data-driven mechanisms to map the impact of context on model outputs, our approach explicitly integrates context to guide knowledge adaptation, which offers a more interpretable and structured way to enhance robot learning efficiency.

\section{Preliminaries and Problem Formulation} \label{sec:problem}
In this section, we introduce the fundamentals of Markov Decision Processes (MDPs) and define the concept of \textit{knowledge} for solving an MDP. Building on these, we formulate the problem of context-aware adaptation, which seeks to enable robots to efficiently learn new tasks by leveraging prior knowledge.

\subsection{Markov Decision Processes and Knowledge} 
An RL task can be modeled by a Markov Decision Process~\cite{wang2022deep} defined as a tuple $\mathbf{T}=\{\mathcal{S},\mathcal{A}, \mathcal{P},\mathcal{R}\}$, where each element represents state space, action space, transition probability measure, and reward function, respectively. 
The objective of an RL agent is to learn a decision-making strategy that selects actions from the action space $\mathcal{A}$ based on current states in state space $\mathcal{S}$, in order to maximize the cumulative reward $\mathcal{R}$.
In this paper, we use a general notation $\mathcal{K}$ to represent the critical knowledge required for such decision-making. Depending on the RL methods employed, $\mathcal{K}$ can take various forms: For policy-based RL~\cite{arulkumaran2017deep}, it represents a policy that directly maps states to actions. For value-based RL~\cite{arulkumaran2017deep}, it represents the value function that estimates the cost-to-go which informs the choice of action; For actor-critic RL~\cite{arulkumaran2017deep}, it combines both the policy (actor) and value function (critic).

\subsection{Problem Formulation}
This paper concerns the efficient adaptation to a new task by leveraging prior knowledge from previously solved tasks.  
Consider a set of $n$ source tasks, denoted as $\mathbb{T} = \{\mathbf{T}_1, \mathbf{T}_2, \dots, \mathbf{T}_n\}$, where each source task $\mathbf{T}_i = \{\mathcal{S}, \mathcal{A}, \mathcal{P}_i, \mathcal{R}\}$ for $i \in \{1, \dots, n\}$ is defined by shared state and action spaces ($\mathcal{S}$, $\mathcal{A}$) and a common reward function ($\mathcal{R}$), but distinct transition probability measures ($\mathcal{P}_i$).
A new unseen target task is represented as $\mathbf{T}_g=\{\mathcal{S},\mathcal{A}, \mathcal{P}_g,\mathcal{R}\}$.
The target task shares the same state space, action space, and reward function as the source tasks, but its transition probability measure $\mathcal{P}_g$ is unknown.

The formulation applies to scenarios with similar objectives, such as robot navigation tasks with a consistent success metric. We can use the same state and action spaces for all scenarios, but variations in robot configurations (e.g., chassis, wheels, tires, etc.) and deployment across various terrain (e.g., geometry, surface material, etc.) will lead to differences in transition probability measures.

We assume the source tasks can be sufficiently trained to acquire the corresponding set of knowledge $\mathbb{K}=\{\mathcal{K}_1,\mathcal{K}_2,\cdots,\mathcal{K}_n\}$. The problem of interest is to develop an efficient algorithm that derives the knowledge $\mathcal{K}_g$ for the target task $\mathbf{T}_g$, by leveraging the prior knowledge $\mathbb{K}$ to accelerate learning and improve performance.

\section{Context-aware Adaptation for Robot Learning (CARoL)}

To explain the rationale behind CARoL, note that in a given MDP, the transition probability $\mathcal{P}_i$ plays a central role in determining the knowledge $\mathcal{K}_i$ required to solve the problem. This makes $\mathcal{P}_i$ a natural choice as the contextual marker for knowledge adaptation: greater similarities in state transitions indicate stronger relevance between the knowledge associated with a source task and that needed for a new target task.
While existing methods often rely on environmental or robot features to implicitly capture this relationship, our approach explicitly uses state transitions as the context for adaptation. This provides a more interpretable and structured method to enhance learning efficiency.

\begin{figure}
    \centering
    \includegraphics[width=0.5\textwidth]{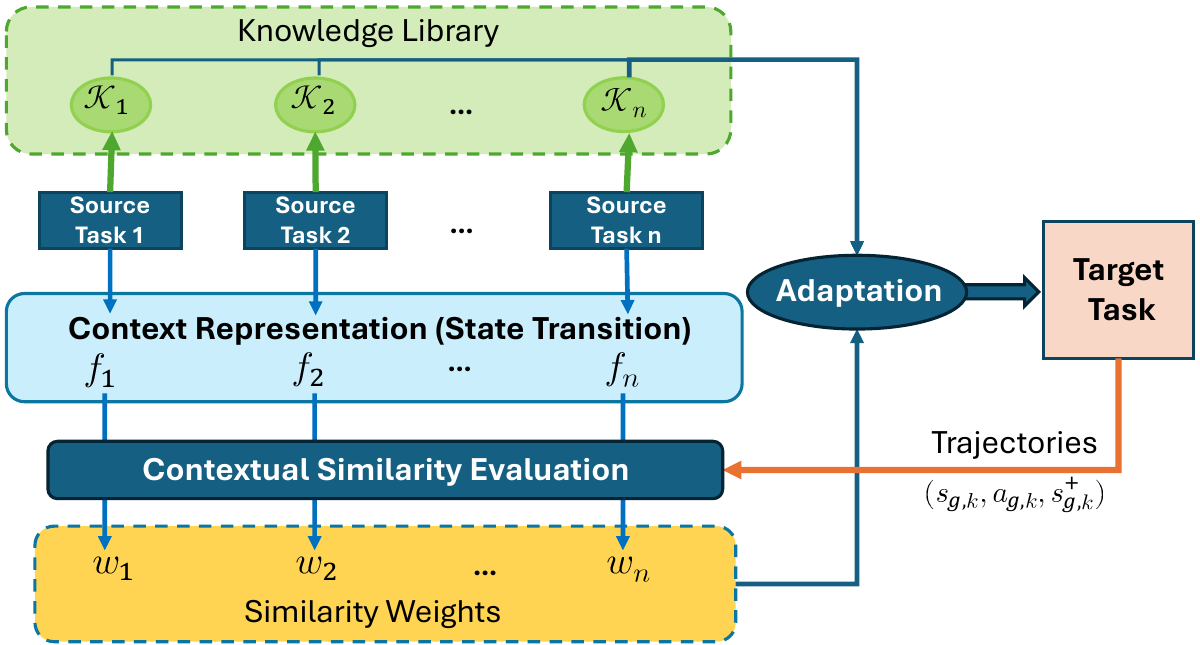}
    \caption{The overview of Context-aware Adaptation for Robot Learning (CARoL) framework. }
    \label{fig:alg_overview}
\end{figure}

In the following, we present details of the proposed CARoL algorithm. 
As illustrated in Fig. \ref{fig:alg_overview}, CARoL consists of three key steps. 
\begin{enumerate}
    \item Acquiring both the context representation and prior knowledge for each source task.
    \item Evaluating the contextual similarities between the target task and source tasks.
    \item Performing contextual knowledge adaptation, which can be applied to policy-based, value-based, or actor-critic RL approaches.
\end{enumerate}

\noindent\textbf{Context Representation.}
For each source task $\mathbf{T}_i$, along with the standard use of RL that acquires the knowledge $\mathcal{K}_i$, we also train an approximation of the transition function to represent the context, denoted by $f_i(s_i,a_i~|\phi_i):\mathcal{S}\times \mathcal{A}\to \mathcal{S}$ for each source task. This function predicts the next state based on the current state-action pair and is parameterized by $\phi_i$. The parameters are updated via gradient descent for the $i$-th task as follows:
\begin{align}
    \label{eqa:gradient}
    \phi_i' = \phi_i - \alpha \nabla_{\phi_i} \mathcal{L}_{\mathbf{T}_i}(\phi_i),
\end{align}
where $\alpha$ is the learning rate and the task collects state-action-next state triplets $\{(s_{i,k}, a_{i, k} ,s^+_{i, k})\}$ sampled from robot trajectories, with $k\in\{1,\cdots,m_i\}$ being the indicator of samples. The loss function is defined as the difference between the next true state from samples and the predicted next state:
\begin{align}
    \mathcal{L}_{\mathbf{T}_i}(\phi_i) = \sum_{k=1}^{m_i} \|f_i(s_{i,k}, a_{i,k}~|\phi_i) - s^+_{i,k}\|^2.
\end{align}

\medskip
\noindent\textbf{Evaluation of Contextual Similarity.}
We use the state transition functions $f_i(\cdot)$, $i\in\{1,\cdots,n\}$ to evaluate the contextual similarity between source tasks and the new target task which quantifies the relevance of their knowledge. To this end, a straightforward approach is to learn a transition function $f_g(\cdot)$ for the new task, then compare it with $f_i(\cdot)$ from source tasks. However, learning $f_g(\cdot)$ for a new task can be computationally expensive. 

To address this, our method uses sampled trajectories $\{(s_{g,k}, a_{g, k} ,s^+_{g, k})\}$, $k\in\{1,\cdots,m_g\}$ from the target task. These samples are applied to each state transition function $f_i(\cdot)$, $i\in\{1,\cdots,n\}$ learned from the source tasks to compute their similarity, that is, for source task $i$:
\begin{align}
    \label{eqa:transition}
    \mathcal{Y}_{i} = \sum_{k=1}^{m_g}\|f_i(s_{g, k}, a_{g, k})- s^+_{g, k}\|^2.
\end{align}
Based on \eqref{eqa:transition}, a smaller value of $\mathcal{Y}_{i}$ indicates a higher contextual similarity between $i$-th source task and the target task. To compute similarities for all source tasks, we regulate these measures into weights as $i\in \{1,\cdots,n\},$
\begin{align}
    \label{eqa:weight}
    w_i = \frac{\exp(-\mathcal{Y}_{i})}{\sum_{j=1}^n \exp(-\mathcal{Y}_{j})}.
\end{align}

\noindent\textbf{Context-aware Adaption.}
The similarity weights $w_i$ enable the robot to prioritize knowledge from the corresponding source tasks. This makes its adaptation to the target task more focused and effective. 
In RL, different approaches such as policy-based, model-based, or actor-critic-based have distinct learning mechanisms. Thus, the adaption processes are also different, as we describe next.

\subsubsection{CARoL for Policy-based RL}
We consider the case where the knowledge $\mathcal{K}_i$ of each source task is a pre-trained policy, denoted by $\pi_i(s):\mathcal{S}\to\mathcal{A}$. 
The contextual similarity weights are $w_i$.

Let the target policy $\pi_g(s|\theta_g)$ be parameterized by $\theta_g$ and assume it interacts with the target environment on-policy at each iteration to generate an action distribution 
$\pi_g(s|\theta_g)$ based on the current state $s \in \mathcal{S}$ and parameter $\theta_g$. Let $\xi_g$ represent the set of trajectories explored by the policy in the target task and each trajectory $\tau \in \xi_g$ is defined as a sequence of state-action pairs.

To incorporate the knowledge from source policies $\pi_i(s)$, we define a learning loss based on the divergence between the action distribution of the target policy and a weighted combination of the action distributions of the source policies:
\begin{align}
    \label{eqa:policy_adaptation}
    \mathcal{L}_{P}=\sum_{\tau\in \xi_g} \sum_{s \in \tau}\sum_{i=1}^n w_i \mathcal{D} (\pi_i(s), \pi_g(s|\theta_g)),
\end{align}
Where $\mathcal{D}(\cdot, \cdot)$ is the Kullback–Leibler (KL) divergence: 
\begin{align}
    \label{eq:kl}
    \mathcal{D} (\pi_i(s), \pi_g(s|\theta_g)) =  \Phi(\frac{\pi_i(s)}{T}) ln \frac{\Phi(\frac{\pi_i(s)}{T})}{\Phi( \pi_g(s|\theta_g)))}.
\end{align}
Here, $\Phi(\cdot)$ is the softmax function and $T$ is the temperature parameter. 
By minimizing the re-weighted loss $\mathcal{L}_{P}$, the parameters of the target policy $\theta_g$ are iteratively updated through gradient-based optimization. The process selectively incorporates knowledge from relevant source policies to optimize the target task, as summarized in Algorithm \ref{Alg_policy}.

\begin{algorithm2e}\label{Alg_policy}
\SetAlgoLined
\caption{CARoL for Policy-based RL}
    \textbf{input} Knowledge as pre-trained source policies $\{\pi_1,  \dots, \pi_n\}$; contextual similarity weights $\{w_1,\cdots,w_n\}$; the target task $\mathbf{T}_g$;\\
    \textbf{initialize} target policy $\pi_g(s|\theta_g)$; a learning rate $\eta$;\\
    \For{iteration = 1 to $K$}{
        Collect trajectories $\xi_g$ by interacting with the target task $\mathbf{T}_g$ using $\pi_g(s|\theta_g)$;\\
            \For{each minibatch of data from $\xi_g$}{
            Compute $\mathcal{L}_{P}$ with equation ~\eqref{eqa:policy_adaptation};\\
            Update: $\theta_g' = \theta_g - \eta \cdot \nabla_{\theta_g} \mathcal{L}_{P}$;\\
            }
    }
\Return{target policy $\pi_g$.}\
\end{algorithm2e}

\subsubsection{CARoL for Value-based RL}
Value-based RL algorithms use value functions to estimate the expected cumulative reward starting from a given state-action pair (or from a state). Value functions are used to guide action selection.
In the following, we consider the knowledge $\mathcal{K}_i$ of each source task to be a pre-trained state-action value function\footnote{Note that a similar mechanism can be easily generalized to the case of state value function $V_i(s)$.} $Q_i(s,a):\mathcal{S}\times\mathcal{A}\to\mathbb{R}$. 
The contextual similarity weights are $w_i$.

Let the target value function be a neural network $Q_g(s,a~|\psi_g)$ parameterized by $\psi_g$, which estimates the value of a state-action pair for the target task. 
Following the general approach in value-based RL algorithms, actions are greedily selected for the target task using an epsilon-greedy strategy, transitioning from early exploration to later exploitation.
During exploitation, we select the action that yields the highest $Q$-value output from the current state in the $Q$-network.
Data collected during exploration and exploitation is stored in a replay buffer for future updates.

Once the replay buffer reaches a sufficient size, we randomly extract a minibatch $\mathcal{B}$ of data. For each data pair in this minibatch, the target value is computed. However, unlike traditional value-based RL algorithms, we do not rely on the inadequately trained target $Q_g$-network for updates. Instead, we leverage the knowledge from the source $Q_i(s,a)$ functions, reweighted by contextual similarities, as follows:
\begin{align}
    \label{eqa:q_next}
    Q_{\text{next}} = \sum_{i=1}^n w_i Q_i(s^+, \arg \max\limits_{a^+} (Q_i(s^+, a^+))),
\end{align}
where $s^+$ is the next state, and $\arg\max$ selects the action by maximizing each source $Q_i$ value function.

Using the Bellman equation with a discount factor $\gamma$, we define the following learning loss based on the difference between the target value function and the value using reweighed source $Q_i$ networks as prior knowledge:
\begin{align}
    \label{eqa:td_loss}
    \mathcal{L}_{Q}= \sum_{s \in \mathcal{B}}\left[\|Q_g(s,a~|\psi_g)-r - \gamma Q_{\text{next}}\|^2\right],
\end{align}
where $\mathcal{B}$ is the minibatch extracted from the replay buffer, and $r$ is the reward for the corresponding state-action pair.

By minimizing the reweighed loss $\mathcal{L}_{Q}$, the parameters of the target value function, $\psi_g$, are iteratively updated by the following Algorithm \ref{Alg_value}. The process summarized in Algorithm \ref{Alg_value} selectively incorporates knowledge from the relevant source value functions to approximate the value function for the target task. To test the learned $Q_g(s,a~|\psi_g)$ in $\mathbf{T}_g$, the greedy method can be used to select actions.

\begin{algorithm2e}[h]\label{Alg_value}
\SetAlgoLined
\caption{CARoL for Value-based RL}
    \textbf{input} Knowledge as pre-trained source Q networks $\{Q_1, Q_2, \dots, Q_n\}$; contextual similarity weights $\{w_1,\cdots,w_n\}$; the target task $\mathbf{T}_g$;\\
    \textbf{initialize} target value function network $Q_g(s, a~|\psi_g)$; a replay buffer $\mathcal{M}$; a learning rate $\eta$\\
    \For{episode = 1 to $K$}{
    Greedy method to choose action $a$ with explorations or $\arg \max\limits_a (Q_g(s, a~|\psi_g))$ in target task $\mathbf{T}_g$ to get next state $s'$ and reward $r$;\\
    Store $(s, a, r, s')$ in $\mathcal{M}$;\\
    \If{the size of $\mathcal{M}$ is sufficient}{
    Randomly sample a minibatch $\mathcal{B}$ from $\mathcal{M}$;\\
    Compute $\mathcal{L}_{Q}$ with equation ~\eqref{eqa:td_loss};\\
    Update: $\psi_g' = \psi_g - \eta \cdot \nabla_{\psi_g} \mathcal{L}_{Q}$;\\
    }
    }
\Return{ target value function $Q_g$.}\
\end{algorithm2e}

\subsubsection{CARoL for Actor-Critic RL}
If the prior knowledge $\mathcal{K}_i$ is in the form of an actor-critic structure, involving both a policy $\pi_i(s)$ (actor) and a value function $Q_i(s,a)$(critic), then both of them can be leveraged to enhance learning in a new task.
Instead of directly combining the methods introduced earlier to simultaneously learn a policy and a value function for the target task, we focus on learning a target actor (i.e., policy) $\pi_g(s \mid \theta_g)$, while adopting a static fusion method to leverage the source critic (value) functions without using a parameterized function approximation. Otherwise, if the actor and critic are updated simultaneously, the method might suffer from issues such as instability and overestimation of $Q$-values, which are common in actor-critic RL~\cite{van2016deep}.

Similar to the policy-based approach outlined in Algorithm \ref{Alg_policy}, we define part of the learning loss to be $\mathcal{L}_P$, which measures the weighted combination of divergences between the target policy's action distribution and source action distributions.


Additionally, recall that in actor-critic methods, the actor $\pi_g$ is trained to select an action $a = \pi_g(s)$ for a given state $s$ that maximizes the value estimated by the critic. Based on this principle, we define a loss from critic as follows:
\begin{align}
    \label{eqa:actor_critic_loss_2}
    \mathcal{L}_{C}=-\sum_{\tau\in \xi_g} \sum_{s \in \tau}\left[\sum_{i=1}^n w_i \cdot Q_i(s,\pi_g(s|\theta_g))\right].
\end{align}
In this loss function, the action for each state $s$ is determined by the current actor $\pi_g$. Thus, $\mathcal{L}_{C}$ evaluates the performance of the target actor using the weighted combination of source critics, prioritizing those that are more contextually similar to the target task. Minimizing $\mathcal{L}_{C}$ helps to iteratively improve the target actor by leveraging the guidance of source critics. 

The combined losses in \eqref{eqa:policy_adaptation} and \eqref{eqa:actor_critic_loss_2} yields:
\begin{align}
    \label{eqa:lossAC}
    \mathcal{L}_{AC}=\mathcal{L}_{P}+\beta\mathcal{L}_{C},
\end{align}
where $\beta\in\mathbb{R}_+$ is an adjustable parameter. 
By minimizing $\mathcal{L}_{AC}$, the target actor $\pi_g$ is informed by both the source actors $\pi_i(s)$ through $\mathcal{L}_P$ as well as the source critics $Q_i(s,a)$ through $\mathcal{L}_C$. The value of $\beta$ determines the importance of each. This process ensures that the target actor effectively integrates knowledge from both the source actors and critics to improve performance on the target task. Algorithm \ref{Alg_AC} summarizes how to use CARoL for actor-critic RL.  

\begin{algorithm2e}\label{Alg_AC}
\SetAlgoLined
\caption{CARoL for Actor-Critic RL}
    \textbf{input} Knowledge as pre-trained source actors $\Pi = \{\pi_1, \pi_2, \dots, \pi_n\}$ and critics $\{Q_1, Q_2, \dots, Q_n\}$; contextual similarity weights $\{w_1,\cdots,w_n\}$; the target task $\mathbf{T}_g$;\\
    \textbf{initialize} target actor $\pi_g(s|\theta_g)$;\\
    \For{iteration = 1 to $K$}{
        Collect trajectories $\xi_g$ by interacting with the target task $\mathbf{T}_g$ using $\pi_g(s|\theta_g)$;\\
            \For{each minibatch of data from $\xi_g$}{
            Compute $\mathcal{L}_{AC}$ with equation ~\eqref{eqa:policy_adaptation} and \eqref{eqa:actor_critic_loss_2};\\
            Update: $\theta_g' = \theta_g - \eta \cdot \nabla_{\theta_g} \mathcal{L}_{AC}$;\\
            }
    }
\Return{target actor $\pi_g$.}\
\end{algorithm2e}

\begin{remark}\label{RM1}
In this section, the introduced CARoL focuses on leveraging prior knowledge to solve a new target task. It omits the aspect of using standard RL approaches to acquire new knowledge from the task. This is because additional learning may not always be necessary when prior knowledge is already sufficient for solving the task.
However, if learning from the target task is required, this can be achieved by adding standard RL loss terms to the functions \eqref{eqa:policy_adaptation}, \eqref{eqa:td_loss}, and \eqref{eqa:lossAC} for policy-based, value-based, and actor-critic methods, respectively. This straightforward extension enables the model to not only adapt prior knowledge but also refine it through direct learning from the target task. An implementation of this, called CARoL+, will be included in the experiments section for comparison.
\end{remark}

\section{Experiments}
In this section, we present both simulated and physical experiments to evaluate the effectiveness of CARoL in leveraging prior knowledge and context awareness to solve new tasks.

\begin{figure}[h]
    \centering
    \includegraphics[width=0.45\textwidth]{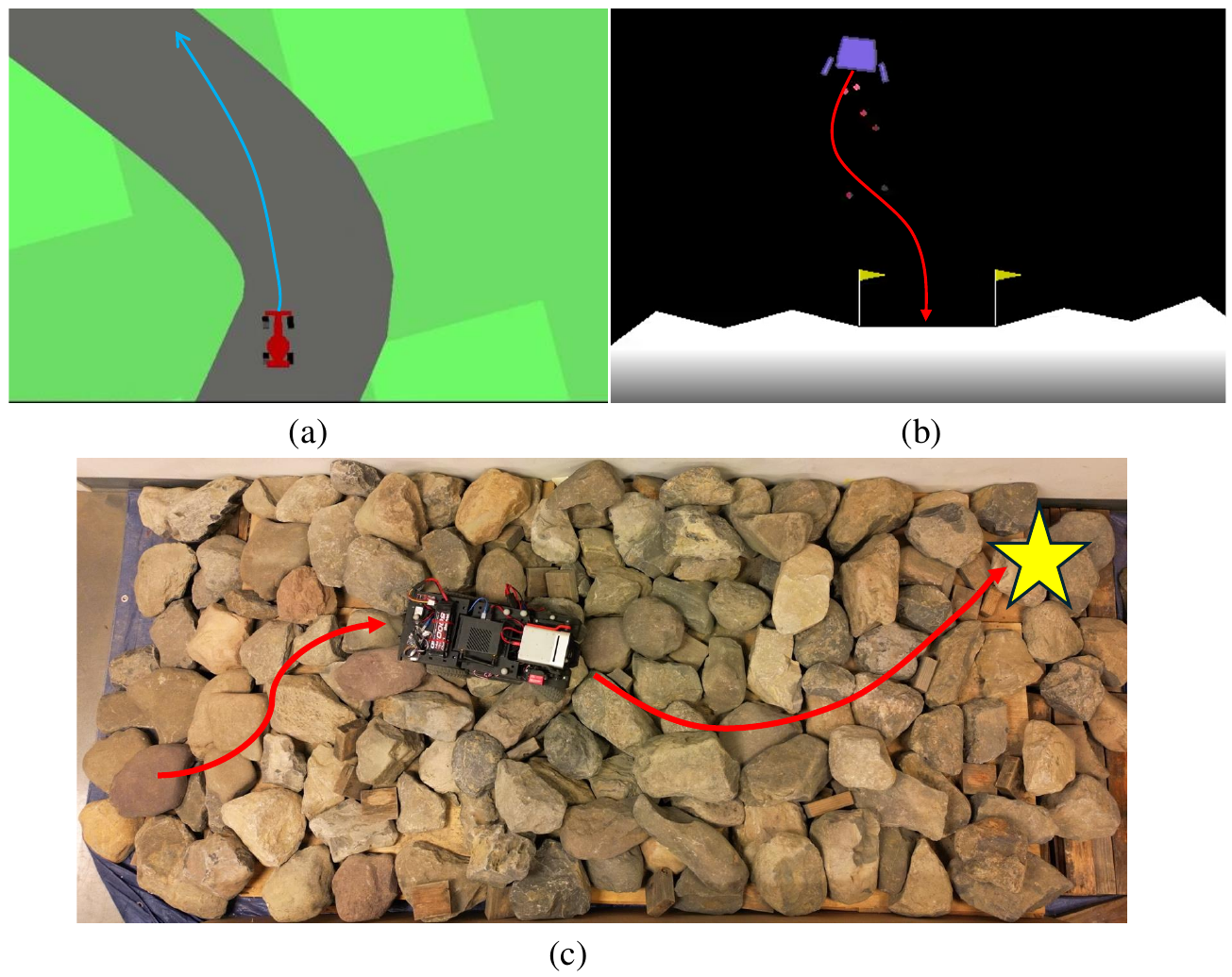}
    \caption{Two OpenAI simulation environments for validating our algorithm: (a) CarRacing environment; (b) LunarLander environment. One physical experiment environment: (c) Ground vehicle off-road navigation.}
    \label{fig:sim_intro}
\end{figure}

To ensure reproducibility, the simulated experiments are conducted using OpenAI's CarRacing and LunarLander tasks, as shown in Fig.~\ref{fig:sim_intro}-(a,b).
To demonstrate CARoL's capabilities in real-world scenarios (Fig. ~\ref{fig:sim_intro}-(c)), we deploy it on a ground vehicle, enabling the rapid and efficient adaptation of policies learned from source tasks to smoothly navigate real-world off-road terrain.
In these setups, CARoL is integrated with various RL algorithms for knowledge adaptation, and its performance is compared against multiple baseline approaches.

\subsection{Simulated Experiment: CarRacing}
\begin{figure*}[h]
    \centering
    \includegraphics[width=0.85\textwidth]{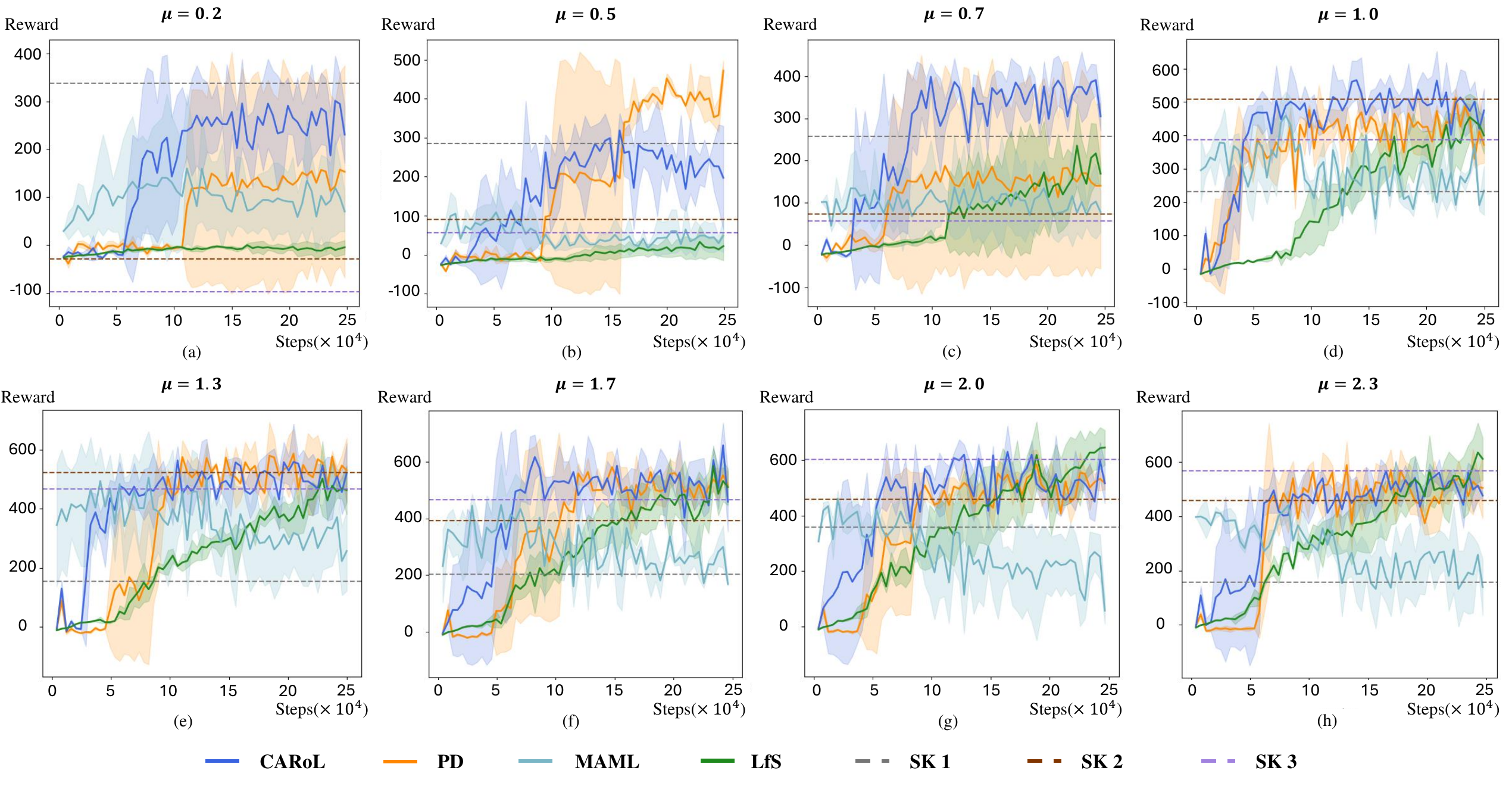}
    \caption{Comparison of cumulative reward for CARoL and baselines under different CarRacing tasks. The x-axis represents the running steps and the y-axis represents the reward. The height of dash lines demonstrates the average accumulative reward of source knowledge (SK) under the task.}
    \label{fig:carracing_reward}
\end{figure*}
\subsubsection{Experiment Setup}
As shown in Fig. \ref{fig:sim_intro}-(a), CarRacing is a continuous control task where an autonomous mobile robot must navigate a procedurally generated racetrack to maximize cumulative rewards. The task involves balancing speed and control to avoid going off-track while efficiently moving forward on the racetrack. 

\begin{itemize} [leftmargin=*]
\item The observation space of the task includes four $96\times 96 \times 3$ RGB images of the top-down view of the car and track (including the current and past three time-steps), as well as the speed and gyroscope readings of the car.

\item The continuous action space consists of three dimensions: steering $[-1, 1]$, acceleration $[0,1]$, and brake $[0,1]$. 

\item The robot receives a reward for visiting new segments of the track and is penalized for each timestep. The episode terminates when the car completes the track or it goes off-track. 
\end{itemize}
In CarRacing, the dynamics of the car depend on track friction. We design three different task environments with low $\mu=0.5$, medium $\mu=1$, and high friction $\mu=2$, as source tasks. Our goal is to quickly obtain a policy in a target task with a new friction level. The friction levels for both source and target tasks are unknown to the robot.

\smallskip
\subsubsection{Implementation of CARoL and other Baselines}

\begin{itemize} [leftmargin=*]
    \item \textbf{CARoL}: In CarRacing, we obtain all source knowledge using PPO (Proximal policy optimization)~\cite{schulman2017proximal}. While PPO has a value function that is used to improve training stability, it is primarily an actor-focused method. The source knowledge is represented as pre-trained source policies. Therefore, our CARoL framework utilizes Algorithm \ref{Alg_policy} for adaptation to the target tasks.


    \item \textbf{Policy Distillation (PD)} \cite{rusu2015policy}: We include PD as a representative method for multi-task reinforcement learning in our baseline. PD is implemented using well-trained source policies as teachers to train a target policy. The key distinction between PD and CARoL lies in their adaptation mechanisms: CARoL leverages similarity weights to enable context-aware adaptation, whereas PD fuses all source policies equally in an indiscriminate manner. 
    

    \item \textbf{Model-Agnostic Meta-Learning (MAML)} \cite{finn2017model}: We include MAML as a representative method for meta-learning in our baseline. The meta-model is trained using all source tasks. Then the meta-model is adapted to the new environment using PPO.
    

    \item \textbf{Learning from Scratch (LfS)}: We incorporate the same method that is used in source knowledge training but adjust the network structure to match that of CARoL, enabling learning from scratch for the target task.

    \item \textbf{Source Knowledge (SK)}: We directly apply source knowledge to different target tasks. The accumulative reward values are evaluated by averaging over 20 episodes.
\end{itemize} 
\textbf{Implementation Details}: 
    For training SKs, we employ a CNN-based actor network, consisting of 6 convolutional layers with increasing channels \{8, 16, 32, 64, 128, 256\}
    and a fully connected layer (100 units). The input of the actor-network is obtained by: first converting the (top-down view) RGB image to a grayscale image, then stacking the current image together with the images of the three previous timesteps. 
    To enable context awareness of CARoL, we learn the state transition function using a 4-layer MLP 
    \{16, 32, 16, 2\}
    with ReLU activations in the hidden layers. This model considers vehicle speed and gyroscope readings as states and steering, acceleration, and braking as actions. 
    CARoL, PD, and LfS each output a policy for the target task with the same architecture: a lighter actor network with 4 convolutional layers 
    \{8, 16, 32, 64\}
    and a fully connected layer (50 units). All models are trained using the Adam optimizer with a fixed learning rate of $1\times 10^{-4}$ and a discount factor of 0.99. The total training timesteps for SK and the inner-loop of source tasks in MAML are set to $6\times 10^6$ timesteps, while the adaptation phase for CARoL, PD, and LfS is limited to $2\times 10^6$ timesteps for faster adaptation and fair comparison. For MAML, we train the meta-policy with a meta-batch size of 3 source tasks, and the adaptation is performed over 5 gradient steps with the same learning rate.

\begin{table}[]
    \caption{Contextual similarities with each source tasks in CarRacing.}
    \label{tab:carracing_weights}
    \centering
    \begin{tabular}{l|ccc}
    \toprule
    & $\mathbf{T}_1$ & $\mathbf{T}_2$ & $\mathbf{T}_3$ \\
    \midrule 
       \text{a: }$\mathbf{\mu=0.2}$  & 1.00 & 0.00 & 0.00 \\
       \midrule 
       \text{b: }$\mathbf{\mu=0.5}$  & 0.98 & 0.02 & 0.00 \\
       \midrule 
       \text{c: }$\mathbf{\mu=0.7}$  & 0.83 & 0.17 & 0.00 \\
       \midrule 
       \text{d: }$\mathbf{\mu=1.0}$   & 0.05 & 0.94 & 0.01 \\
       \midrule 
       \text{e: }$\mathbf{\mu=1.3}$   & 0.10 & 0.64 & 0.26 \\
       \midrule 
       \text{f: }$\mathbf{\mu=1.7}$  & 0.06 & 0.15 & 0.79 \\
       \midrule 
       \text{g: }$\mathbf{\mu=2.0}$   & 0.06 & 0.11 & 0.83 \\
       \midrule 
       \text{h: }$\mathbf{\mu=2.3}$   & 0.08 & 0.13 & 0.79 \\
       \bottomrule
    \end{tabular}
\end{table}

\smallskip
\subsubsection{Experiment Results and Analysis}

Table \ref{tab:carracing_weights} presents the contextual similarity weights learned for the three source tasks in different target tasks under the CarRacing simulated experiment. We observe that the contextual similarity weight assigned to a source task for a given target task is generally negatively correlated with their friction difference. In other words, the friction of source tasks closer to the target task tends to receive higher similarity weights, and vice versa. 
In particular, the tasks with $\mu=0.5$, $\mu=1$, and $\mu=2$ correspond to scenarios where the target task is the same as one of the source tasks. These serve as a sanity check for the algorithm's ability to correctly identify contextual similarity. The results confirm the effectiveness of our approach, as the corresponding source task consistently receives the highest similarity weight.

In terms of training details, Fig.~\ref{fig:carracing_reward} presents the reward curves, where the x-axis represents the number of training steps, and the y-axis represents the reward value. 
We observe that each SK performs well on its corresponding task, indicating that the SK has been well-trained. 
In all cases, CARoL demonstrates a faster convergence speed compared to other methods. This verifies CARoL's capability to efficiently leverage existing knowledge and quickly adapt to the target task. In terms of the final reward, CARoL generally achieves the best performance. Specifically, when the target environment matches one of the source environments, as in subfigures (b, d, g), CARoL produces results comparable to the well-trained source knowledge (SK) policies.
When compared to PD, which fuses SK policies in an indiscriminate manner, CARoL can prioritize relevant SKs, leading to better performances in (a, c). Additionally, CARoL exhibits a more stable training process, whereas PD suffers from significant oscillations, as observed in (a, c) and middle parts of (b, e, f, g).

The LfS uses the same parameter settings as CARoL. LfS struggles to converge in most tasks within the given training steps, whereas CARoL adapts significantly faster. Furthermore, LfS exhibits a clear trend of training inefficiency, where more challenging tasks (e.g., environments with low friction) take longer to converge. This is because, in lower friction settings, vehicles are more prone to going off-track during exploration. In contrast, CARoL is not affected by this issue. 
Regarding MAML, despite extensive testing meta-training and adaptation, its performance remains poor. This may be due to an undesirable meta-model that is difficult to generalize effectively to new environments (If trained sufficiently long, the MAML rewards would start to increase and match that of LfS.). Nevertheless, regardless of the final performance, meta-training is computationally expensive, whereas CARoL does not need such a process and can leverage SK directly.

\subsection{Simulated Experiment: LunarLander}
\subsubsection{Experiment Setup}
As shown in Fig. \ref{fig:sim_intro}-(b) LunarLander is a control task where an autonomous lander robot must safely land on a designated landing pad in a simulated lunar environment. The robot must balance thrust and control to achieve a smooth landing while minimizing fuel consumption. 
\begin{itemize} [leftmargin=*]
    \item The observation space is an 8-dimensional vector representing the lander’s position, velocity, angle, and contact sensors for the landing legs.

    \item The continuous action space determines the throttle percentage of the vertical main engine $[-1, 1]$, and the throttle of the lateral boosters $[-1, 1]$. 

    \item Positive rewards are given for a smooth landing, and penalties are applied for crashes, excessive fuel consumption, and staying off the designated landing pad. 
\end{itemize}

\begin{figure}[h]
    \centering
    \includegraphics[width=0.3\textwidth]{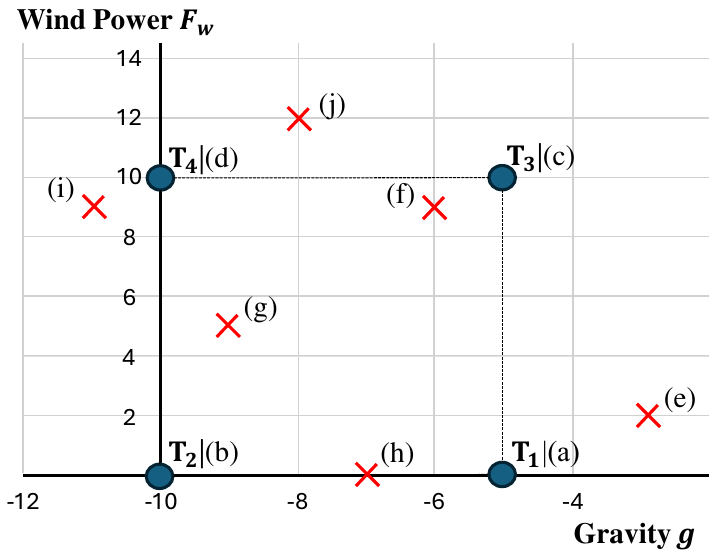}
    \caption{Task configurations in Lunarlander. In the plot, circles represent the source tasks ($\mathbf{T}_1$-$\mathbf{T}_4$); crosses represent the target tasks (a-j). 
    This figure intuitively visualizes the similarity relationships between the source tasks and target tasks. The environment configurations (for both source and target tasks) are \textbf{unknown} to the robot. }
    \label{fig:lunarlander_config}
\end{figure}

The LunarLander environment allows adjustments of gravity and wind power, based on which we can create different source and target tasks. 
Different gravity levels, $g \in [-12, -2]$,
affect the vertical force on the lander robot, while varying wind power, $F_w \in [0, 10]$, impacts the horizontal force\footnote{In our experiments, wind power is modeled as a time-invariant fixed value to emphasize its impact on state transitions. This differs from the standard environment, where wind power is randomized at each time step.}.
The configuration of source tasks and target tasks for this experiment is visualized in Fig.~\ref{fig:lunarlander_config}. The horizontal axis represents gravity, the vertical axis represents wind power. They together expand a 2-D configuration space. The blue dots indicate the source tasks, which are defined as follows: $(-5, 0)$; $(-10, 0)$; $(-5, 10)$; $(-10, 10)$. The red crosses represent the target tasks used for testing. These target tasks are located above, to the left, to the right, and inside a square formed by the source tasks over the configuration space. 
Note that the environment configurations for both source and target tasks are unknown to the robot. Using context awareness, the robot will derive a unique prioritization of source knowledge for each target task.

\smallskip
\subsubsection{Implementation of CARoL and other Baselines}
\begin{itemize} [leftmargin=*]
    \item \textbf{CARoL}: In LunarLander, we obtain all source knowledge using TD3 (Twin Delayed Deep Deterministic)\cite{fujimoto2018addressing}. 
    The source knowledge is represented as pre-trained source policies and state-action value functions. Therefore, our CARoL utilize Algorithm \ref{Alg_AC} for adaptation on the target tasks. Here, we consider the actor and critic with equal importance, i.e., $\beta=1$
    

    \item \textbf{CARoL+}: The implementation of CARoL+ incorporates CARoL's loss function~\eqref{eqa:lossAC} and a standard TD3 learning loss. As explained in Remark~\ref{RM1}, this facilitates knowledge adaptation not only through CARoL but also via reward-based learning from the environment.


    \item \textbf{Other baselines that have been introduced in CarRacing}\footnote{In this part, MAML is excluded due to its poor performance in the CarRacing and the excessively long training time of the meta-policy.}: For all other baselines, we use TD3 to obtain the source knowledge, and their network architectures are kept consistent with the target actor initialized in CARoL. 
\end{itemize}
\textbf{Implementation Details}: 
    For training SKs, we use a fully connected actor network, consisting of the architecture with a 4-layer MLP with layer sizes
    \{96, 192, 96, 2\}
    and ReLU activations in the hidden layers, followed by a tanh activation in the output layer. 
    The critic network uses the same architecture. 
    For context awareness of CARoL, we use the same network in CarRacing for state transition functions. The input consists of observed state and the executed action, and its output is the next observed state.
    For CARoL, CARoL+, PD, and LfS, we use the same but lighter actor and critic networks, i.e., 
    \{48, 96, 48, 2\}.
    All methods are trained using AdamW with a fixed learning rate of $4 \times 10^{-4}$ and a discount factor $0.98$. The number of total training episodes for SKs is 15,000 episodes with a maximum of 1,000 steps per episode while others are 10,000 episodes with a maximum of 500 steps per episode. 




\begin{table}[]
    \caption{Contextual similarities with each source task in LunarLander.}
    \label{tab:lunarlander_weights}
    \centering
    \begin{tabular}{l|cccc}
    \toprule
         & $\mathbf{T}_1$ & $\mathbf{T}_2$ & $\mathbf{T}_3$ & $\mathbf{T}_4$\\
         \midrule 
       \text{a: }\textbf{(-5, 0)}  & 0.95 & 0.05 & 0.00 & 0.00\\
       \midrule 
       \text{b: }\textbf{(-10, 0)}  & 0.01 & 0.98 & 0.00 & 0.01\\
       \midrule 
       \text{c: }\textbf{(-5, 10)} & 0.06 & 0.01 & 0.91 & 0.02\\
       \midrule 
       \text{d: }\textbf{(-10, 10)}   & 0.00 & 0.01 & 0.01 & 0.98\\
       \midrule 
       \text{e: }\textbf{(-3, 2)}   & 0.62 & 0.09 & 0.28 & 0.01\\
       \midrule 
       \text{f: }\textbf{(-6, 9)}   & 0.09 & 0.00 & 0.81 & 0.10\\
       \midrule 
       \text{g: }\textbf{(-9, 5)}   & 0.19 & 0.37 & 0.18 & 0.26\\
       \midrule 
       \text{h: }\textbf{(-7, 0)}   & 0.35 & 0.34 & 0.22 & 0.09\\
       \midrule 
       \text{i: }\textbf{(-11, 9)}   & 0.01 & 0.12 & 0.10 & 0.77\\
       \midrule 
       \text{j: }\textbf{(-8, 12)}   & 0.13 & 0.17 & 0.29 & 0.41\\
       \bottomrule
    \end{tabular}
\end{table}

\begin{figure*}
    \centering
    \includegraphics[width=0.95\textwidth]{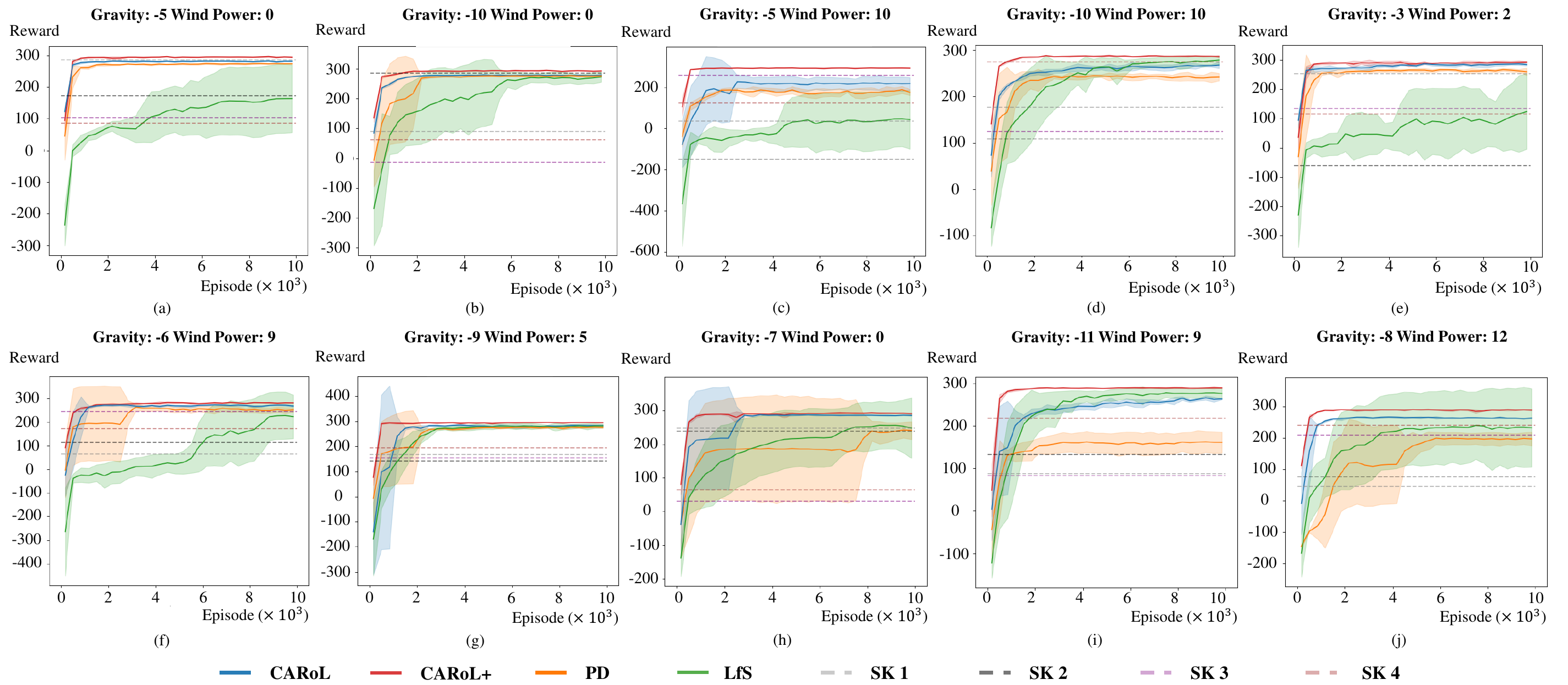}
    \caption{Comparison of cumulative reward for CARoL and baselines under different LunarLander tasks. The x-axis represents the running episodes and the y-axis represents the reward. The height of dash lines demonstrates the average accumulative reward of source knowledge (SK) under the task.}
    \label{fig:lunarlander_reward}
\end{figure*}

\smallskip
\subsubsection{Experiment Results and Analysis}

Table \ref{tab:lunarlander_weights} presents the contextual similarity weights learned for four source tasks in different target tasks under the LunarLander simulated experiment. Similar to the CarRacing, we observe that the weight is negatively correlated with their geometric distance in the figure. We also provide a sanity check here as the first four rows of the table correspond to the same scenarios as source tasks. We can observe a similar pattern to CarRacing. 
Meanwhile, by reading the exact values in the table, we also note that the relationship between weights and distances in Fig.~\ref{fig:lunarlander_config} is not strictly proportional, because the impact from environmental configurations to context (state transition) is nonlinear. For instance, different combinations of gravity and wind power may lead to similar state transitions. Then, the contextual similarity allows the corresponding prior knowledge to be effectively adapted.

In terms of training details, Fig.~\ref{fig:lunarlander_reward} shows the reward curves, where the x-axis represents the number of learning episodes, and the y-axis represents the reward value. CARoL shows faster and higher reward convergence compared to the baselines. 
Subfigures (a-d) correspond to scenarios where the target task is identical to one of the source tasks. In these cases, we observe that each source policy performs best with the highest evaluation reward on its respective source task. 
By prioritizing the most contextually similar knowledge, CARoL achieves comparable accumulated rewards, as seen in (a), (b), and (d). 
For (c), the slight gap of CARoL can be explained by the third row of Table~\ref{tab:lunarlander_weights}, where the weight assigned to $\mathbf{T}_3$ is not as close to 1 as in other cases. By adapting more knowledge from less relevant source tasks, the performance slightly drops. 
Across all other target tasks, where the target is different from any source task, we observe that the average reward of any single source policy is consistently lower than that of CARoL. This is because the source policies lack experience with the target task’s specific contexts. CARoL, on the other hand, effectively combines source knowledge based on contextual similarity, allowing it to achieve better performance.

In comparison with baselines, similar to the observations in CarRacing, we see that PD exhibits a consistently lower reward convergence than CARoL across most target tasks. This is because PD focuses solely on knowledge fusion without considering contextual similarity for new tasks. 
Both PD and CARoL exhibit some training-phase oscillations in the early stages when initially incorporating knowledge from multiple source policies. Nevertheless, these oscillations diminish as the adaption process progresses. Comparing LfS to CARoL, we observe that LfS demonstrates a significantly slower convergence speed, particularly for (f) and (h), and generally achieves lower final rewards.
This is because Lfs has a same network size as CARoL but cannot leverage prior knowledge. Thus, the training is data-demanding and the convergence is difficult. 
It is worth noting that in target tasks (i) and some trials of (j),  LfS achieves higher rewards than CARoL. This is because the source task may not provide sufficient useful knowledge to the target tasks, as suggested by Fig.~\ref{fig:lunarlander_config} that these target tasks use parameters outside the square of the source tasks. In these cases, LfS benefits from learning new knowledge from the new task and thus outperforms CARoL.

However, this issue can be addressed by CARoL+. CARoL+ integrates interaction-based learning during adaptation, allowing it to achieve the highest reward across all cases. This compensates for CARoL’s potential limitations in final convergence values compared to LfS in certain cases. Additionally, CARoL+ retains CARoL’s advantage of fast convergence while also being the most stable one during the convergence process.

\subsection{Real World Experiment: Ground vehicle Off-road Navigation}
We apply CARoL to an off-road mobility problem with a physical robot. As shown in Fig. \ref{fig:sim_intro}-(c), the goal is to enable the robot to autonomously and successfully navigate from the start to the goal. RL for real-world off-road mobility is particularly inefficient due to the prohibitive cost of collecting physical vehicle-terrain interaction data. Therefore, in this experiment, we aim to apply CARoL to adapt policies learned in simulation to the real world. 

\subsubsection{Experiment Setup}
We leverage a high-fidelity multi-physics simulator,  Chrono~\cite{tasora2016chrono}, to train two different off-road navigation policies: SK1 for concrete terrain and SK2 for grass terrain. We generate off-road terrain with undulating topographies and vary the friction coefficient between the vehicle tires and the underlying terrain, i.e., 0.9 and 0.4. 
Here, different friction levels influence traversability. For example, for the same slope, friction determines whether the robot can climb it or must bypass it.
In the real world, we construct an off-road mobility testbed with hundreds of rocks and boulders. We posit the friction coefficient of the real-world rocks and boulders to be between 0.57 and 0.73, necessitating adaption from the two SKs into a new policy.

\begin{itemize} [leftmargin=*]
\item The observation space encompasses a 16-dimensional terrain encoding (derived from elevation footprints via a SWAE~\cite{kolouri2018sliced} architecture), normalized heading error, and current speed. 

\item The action space consists of continuous speed and steering commands.

\item The robot receives rewards for moving towards the goal and penalties for large roll and pitch angles, which are indicators of the robot’s stability during navigation.
\end{itemize}

\subsubsection{Implementation of CARoL}
In the simulators, similar to CarRacing, we use PPO to train two source knowledge policies from scratch and train state transition functions for each environment as context representation.
To implement CARoL in the real-world environment, we first compute the contextual similarity. To achieve this, we manually control the vehicle to randomly navigate 20 trials on the real-world off-road terrain to collect trajectory data, instead of relying on random exploration, for real-world sample efficiency.
The contextual similarity is quantified by comparing the prediction errors of the source task transition functions over trajectories collected in the real-world environment.

Based on the obtained contextual similarity and source task policies, we employ CARoL-Algorithm \ref{Alg_policy} to adapt to the real world. Since the adaptation is on-policy, we iteratively collect trajectories using the current policy and update the policy with the collected data. 
The policy model is implemented as a 3-layer fully-connected network with hidden layers
$\{32, 32, 2\}$ using ReLU activations. we use the Adam optimizer with a fixed learning rate of $1\times 10^{-3}$. The discount factor is set to 0.99. 
We perform a total of 10 iterations with 8 navigation trials each (approximately 130 minutes in total), during which the vehicle gradually improve from initially failing (veering off the experimental field, getting stuck halfway, or flipping over) to successfully reaching the goal. 
In the final two iterations, the vehicle successfully reaches the goal in every episode, indicating a good adaptation performance. We then compare this converged model with the source knowledge policies.
We do not implement baseline methods due to the prohibitive cost of collecting extensive physical vehicle-terrain interaction data. For the same reason, we do not employ CARoL+.

\subsubsection{Experiment Results and Analysis}
In the comparison experiments, each model undergoes five trials from the start to the goal. The source knowledge refers to the model trained in the Chrono simulator, which is directly deployed on the robot. The evaluation metrics for this experiment, aligned with the RL rewards, include success rate, roll, and pitch. While traveling time is not evaluated, it is provided in the table as a reference.

As shown in the results Table \ref{tab:physical_results}, CARoL achieves the highest success rate, and the robot demonstrates the most stable navigation under CARoL’s adapted model. Although CARoL takes the longest time to reach the goal, this is expected, as we do not impose any reward or penalty on travel time in this task. The longer duration might suggests that the robot is carefully selecting paths based on traversability.
The primary objective is to ensure that the robot autonomously reaches the goal successfully, and CARoL’s high stability in terms of roll and pitch plays a crucial role in achieving this outcome. SK2, trained on the low friction grass terrain, achieves the shortest traveling time thanks to the increased friction of the real-world rocks and boulders.

\begin{table}
\caption{Physical Experiment Results.}
\centering
\resizebox{\columnwidth}{!}{%
\small
\setlength{\tabcolsep}{4pt}
\begin{tabular}{c|c|cc|ccccc}
\toprule
Method & {Success $\uparrow$} &  {Roll $\downarrow$} & {Pitch $\downarrow$} & {Time (no penalty)} \\
\midrule
CARoL & {\textbf{5/5}} & \textbf{{4.08}\textdegree$\pm$4.65\textdegree} & \textbf{{8.55\textdegree$\pm$5.13}\textdegree} &{16.88s$\pm$1.97s} \\
SK1 & 3/5 & {5.44}\textdegree$\pm$7.13\textdegree & {10.35}\textdegree$\pm$5.29\textdegree & {11.78s$\pm$1.44s}\\
SK2 & 2/5 & {5.75}\textdegree$\pm$5.00\textdegree & {8.61}\textdegree$\pm$5.12\textdegree & \textbf{8.55s$\pm$0.67s}\\
\bottomrule
\end{tabular}%
}
\label{tab:physical_results}
\end{table}

\section{Limitations}
CARoL's current limitations include its reliance on asynchronous processing of context awareness and adaptation. Specifically, the adaption is performed after the calculation of contextual weights, thus, the current CARoL cannot adapt to dynamic contexts in real-time. 
A potential solution is to simultaneously compute similarity weights and adapt source knowledge in real time during the deployment of the RL policy in the target environment.
Besides, CARoL heavily depends on the quality of source knowledge. Introducing an additional mechanism to evaluate the quality of source knowledge could address this limitation. Such a mechanism would enable CARoL to prioritize tasks not only based on contextual similarity but also on the reliability of the source knowledge. This would result in more robust and effective task adaptation.


\section{Conclusion}
In this paper, we presented CARoL. CARoL is a framework that first evaluates contextual similarities between source and target tasks based on their state transitions. Given the prior knowledge from source tasks, CARoL then performs contextual knowledge adaptation for target tasks.
A key advantage of CARoL is its broad applicability to policy-based, value-based, and actor-critic RL methods. Experimental results demonstrated that our algorithm effectively adapts to new tasks. We validated CARoL’s performance in two open-source simulated environments and on a physical robot platform, where CARoL successfully transferred prior knowledge learned in simulation to real-world off-road environments.
For future work, we aim to integrate context similarity evaluation synchronously into the adaptation process. Additionally, we plan to extend CARoL beyond single-robot adaptation to multi-agent environments, leveraging multi-agent coordination to achieve more effective context representation for task adaptation.

\bibliographystyle{plainnat}
\bibliography{references}
\end{document}